\documentclass[10pt,twocolumn,letterpaper]{article}

\usepackage[pagenumbers]{cvpr}      

\usepackage[dvipsnames]{xcolor}

\definecolor{cvprblue}{rgb}{0.21,0.49,0.74}
\usepackage[pagebackref,breaklinks,colorlinks,citecolor=cvprblue]{hyperref}
\usepackage{multirow}
\usepackage{multicol}
\usepackage{booktabs}
\usepackage{tabularx}
\usepackage{graphicx}
\usepackage{amsmath}
\usepackage{amssymb}
\usepackage{booktabs}
\usepackage{cite}
\usepackage{color}
\usepackage[linesnumbered,ruled,vlined]{algorithm2e}

\def\paperID{257} 
\def\confName{CVPR}
\def\confYear{2024}

\begin{document}

\title{
\texttt{ZONE}: Zero-Shot Instruction-Guided Local Editing
}

\author{
    Shanglin Li{$^{1}$}\thanks{These authors contributed equally.}, Bohan Zeng{$^1$}\footnotemark[1], Yutang Feng{$^1$}\footnotemark[1], Sicheng Gao{$^{1}$} Xiuhui Liu{$^{1}$}, Jiaming Liu{$^{2}$}, \\ Lin Li{$^{2}$}, Xu Tang{$^{2}$}, Yao Hu{$^{2}$}, Jianzhuang Liu{$^4$}, Baochang Zhang{$^{1,3,5}$}\thanks{Corresponding author: bczhang@buaa.edu.cn} \\
    \textsuperscript{\rm 1}Beihang University 
    \textsuperscript{\rm 2}Xiaohongshu Inc
    \textsuperscript{\rm 3}Nanchang Institute of Technology, China   \\
    \textsuperscript{\rm 4}Shenzhen Institute of Advanced Technology, China 
    \textsuperscript{\rm 5}Zhongguancun Laboratory, China \\
}
\maketitle

\begin{abstract}
Recent advances in vision-language models like Stable Diffusion have shown remarkable power in creative image synthesis and editing.
However, most existing text-to-image editing methods encounter two obstacles: First, the text prompt needs to be carefully crafted to achieve good results, which is not intuitive or user-friendly. 
Second, they are insensitive to local edits and can irreversibly affect non-edited regions, leaving obvious editing traces. 
To tackle these problems, we propose a Zero-shot instructiON-guided local image Editing approach, termed \texttt{ZONE}. 
We first convert the editing intent from the user-provided instruction (e.g., ``make his tie blue") into specific image editing regions through InstructPix2Pix. We then propose a Region-IoU scheme for precise image layer extraction from an off-the-shelf segment model. We 
further develop an edge smoother based on FFT for seamless blending between the layer and the image.
Our method allows for arbitrary manipulation of a specific region with a single instruction while preserving the rest. Extensive experiments demonstrate that our \texttt{ZONE} achieves remarkable local editing results and user-friendliness, outperforming state-of-the-art methods. Code is available at \url{https://github.com/lsl001006/ZONE}.
\end{abstract}

\section{Introduction}
\label{sec:intro}




Large-scale vision-language models, such as Stable Diffusion \cite{rombach2022high}, DALL·E 2 \cite{ramesh2022hierarchical}, and Imagen \cite{saharia2022photorealistic}, have revolutionized text-guided image editing by bridging the gap between natural language and image content. Trained on vast visual and textual data, these methods harness generative power to manipulate appearance and style in natural images, offering a wide array of possibilities for enhancing and manipulating images in domains such as photography, advertising, e-commerce, and social media. These advancements have opened up new possibilities for text-guided image editing, making it increasingly important in various applications.



State-of-the-art (SOTA) image generative techniques \cite{ramesh2022hierarchical, rombach2022high, yu2022scaling, nichol2022glide} predominantly concentrate on stylization, where the desired appearance is determined by a reference image or textual description, often leading to global image alterations \cite{kwon2022clipstyler, kim2022diffusionclip, sohn2023styledrop}. However, these methods often lack straightforward local editing capabilities, 
and the precise localization of these edits typically needs additional input guidance, such as segmentation masks \cite{avrahami2022blended, lugmayr2022repaint, goel2023pairdiffusion}, making text-driven editing cumbersome and potentially limiting its scope. Recent description-guided works\footnote{In this paper, we call them description-guided diffusion models.} like Prompt-to-Prompt \cite{hertz2022prompt}, DiffEdit \cite{couairon2022diffedit}, and Text2LIVE \cite{bar2022text2live} make noteworthy contributions to mask-free local edits, but they either require complex textual descriptions (\emph{e.g.,} Prompt-to-Prompt requires word-to-word alignment between the source image caption and the edited image caption, and DiffEdit uses query and reference prompts) or need to specify the edited object (\emph{e.g.,} Text2LIVE asks for multiple prompts), which are not user friendly. 
Instruction-guided editing methods\footnote{In this paper, we call them instruction-guided diffusion models.} \cite{el2019tell, zhang2021text, brooks2023instructpix2pix, zhang2023magicbrush} present more elegant characteristics in this regard. They eliminate the need for image-anchored descriptions, requiring only descriptions of the desired edits (\emph{e.g.}, ``make it snowy"), which facilitates concise and intuitive expression. However, these methods suffer from the over-edit problem, potentially distorting high-frequency details in non-edited regions (see Fig.~\ref{fig:teaser}~(b)).

To tackle these problems, we propose \texttt{ZONE}, a \textbf{Z}ero-shot instructi\textbf{ON}-guided local image \textbf{E}diting approach. \texttt{ZONE} provides a more flexible and creative way to manipulate real images with layers. 

Specifically, we leverage the pretrained instruction-guided model, InstructPix2Pix (IP2P) \cite{brooks2023instructpix2pix}, for image editing.
By exploring the attention mechanism of IP2P, we uncover the implicit associations between the editing locations and user-provided instructions in instruction-guided models. This allows us to identify the locations of the edited objects in instructions without the need for extra specification
(\emph{e.g.}, Stable Diffusion-based methods have to specify the tokens of the objects to edit). 
We further enhance this capability by proposing a Region-IoU scheme in conjunction with SAM \cite{kirillov2023segment}, ensuring the mask refinement of the edited image layer. Our \texttt{ZONE} allows arbitrary image editing actions like ``add", ``remove", and ``change", all accomplished with intuitive instructions. Additionally, \texttt{ZONE} supports multi-turn local editing without affecting non-edited regions, empowering high-fidelity local editing without any training or fine-tuning. Comprehensive experiments and user studies demonstrate that \texttt{ZONE} achieves remarkable results and user-friendliness in local image editing, outperforming existing SOTA methods.

To summarize, we make the following key contributions:

\begin{itemize}

    \item We propose \texttt{ZONE}, a zero-shot image local editing method that enables users to edit localized regions of both real and synthetic images with simple instructions. \texttt{ZONE} preserves non-edited regions without loss and allows arbitrary manipulation of edited image layers.

    \item We reveal and exploit the different attention mechanisms between IP2P and Stable Diffusion when processing user instructions for image editing, with intuitive visual comparisons.

    \item We present a novel Region-IoU scheme and incorporate it with SAM for effective edited region refinement, and introduce a Fourier transform-based edge smoother to reduce the artifacts when compositing the image layers.

    \item Comprehensive experiments and user studies demonstrate that \texttt{ZONE} achieves high-fidelity local editing results without any auxiliary prompts, outperforming SOTA methods in photorealism and content preservation.
    
\end{itemize}

\vspace{-1.5mm}

\section{Related Work}
\label{sec:rel}






\subsection{Generative Models for Image Manipulation}
Image manipulation is a fundamental process within the realm of computer vision, involving altering images with the aid of additional conditions like textual prompts, labels, masks, or reference images. Two mainstream editing methods include Generative Adversarial Networks (GANs) and Diffusion Models (DMs). Typical image manipulation tasks comprise image-to-image translation \cite{isola2017image, choi2018stargan, zhu2017unpaired, kim2017learning, saharia2022palette, sohn2023styledrop, wang2022pretraining, duan2023tuningfree}, super-resolution \cite{ledig2017photo, kang2023gigagan, yue2023resshift, gao2023implicit}, inpainting \cite{pathak2016context, iizuka2017globally, lugmayr2022repaint, rombach2022high}, colorization \cite{blanch2019end, Nazeri_2018, lin2023diffcolor, wang2023unsupervised}, and more. Although GAN-based methods excel when dealing with carefully curated data, they struggle with extensive and heterogeneous datasets \cite{karras2019stylebased, karras2020analyzing, mokady2022selfdistilled}. To enhance generative expressiveness, \cite{ho2020denoising, song2020denoising, song2019generative, ho2022cascaded, rombach2022high, zeng2023controllable, zeng2023ipdreamer} utilize DMs to achieve high-quality generation over diverse datasets.
Recent research has yielded promising generation outcomes through the training or fine-tuning of large-scale text-to-image models \cite{huang2024diffusion, brooks2023instructpix2pix, kawar2023imagic, yu2022scaling, meng2021sdedit, ramesh2022hierarchical, saharia2022photorealistic, nichol2022glide}, as well as by harnessing CLIP \cite{radford2021learning} embeddings to guide image manipulation using textual prompts \cite{crowson2022vqgan, kim2022diffusionclip, kwon2022clipstyler}. Some prior works \cite{avrahami2022blended, avrahami2023blended, hertz2022prompt, parmar2023zero, tumanyan2023plug} also demonstrate the zero-shot editing capability of pretrained DMs. Similarly, our method extensively exploits a pretrained DM's generative capability to facilitate diverse and stylized image editing. However, we uniquely explore the implicit relationship between the DM's editing regions during generation and the whole user instructions, enabling fine-grained layer-specific positioning.

\subsection{Localized Image Editing}

Several recent works have made attempts at localized image editing.
Blend Diffusion \cite{avrahami2022blended} proposes a mask-guided method by blending edited regions with the other parts of the image at different noise levels along the diffusion process.
Text2LIVE \cite{bar2022text2live} introduces an RGBA layer generation approach with a CLIP-supervised generator for performing edits of objects in real images and videos.  
Prompt-to-Prompt \cite{hertz2022prompt} controls the spatial layouts of the image corresponding to the words in the prompt through cross-attention modification, enabling local edits by modifying textual prompts. Pix2Pix-Zero \cite{parmar2023zero} preserves the structure of the original image with cross-attention guidance and applies an edit-direction embedding to make changes to localized objects.
Instruction-based editing methods like IP2P \cite{brooks2023instructpix2pix} and MagicBrush \cite{zhang2023magicbrush} are trained or finetuned on triplet datasets to realize intuitive high-quality image editing based on user-provided instructions.
PAIR-diffusion \cite{goel2023pairdiffusion} allows editing the structure and appearance of each masked part in the original image independently.
While these methods produce impressive results within their specific applications, they compromise on local image editing: instruction-guided methods \cite{brooks2023instructpix2pix, zhang2023magicbrush} and attention-based modifications \cite{hertz2022prompt, parmar2023zero} introduce artifacts to non-edited regions, mask-based methods \cite{goel2023pairdiffusion, avrahami2022blended} add complexity to user interactions, and CLIP-based methods \cite{parmar2023zero, bar2022text2live} sacrifice the flexibility of natural language editing. In contrast, our \texttt{ZONE} requires only a single instruction to achieve high-fidelity local image editing with an image layer.



\subsection{Instruction-Guided Editing}
Despite the significant progress of text-to-image models, most require detailed textual descriptions \cite{nichol2022glide, rombach2022high, ramesh2022hierarchical,ruiz2023dreambooth, saharia2022photorealistic} to convey the desired image content, often falling short of user expectations for image editing. In contrast, direct instruction-guided modifications of target regions/attributes offer a more intuitive and convenient approach, such as ``make the girl smile" and ``give him a ball." Recent advancements in instruction-guided editing and generation \cite{el2019tell, zhang2021text, ouyang2022training, zhang2023hive, brooks2023instructpix2pix, zhang2023magicbrush} have made notable progress. For instance, IP2P \cite{brooks2023instructpix2pix} employs GPT-3 \cite{brown2020language} and Prompt-to-Prompt \cite{hertz2022prompt} to synthesize an instruction-editing dataset, utilizes a pretrained Stable Diffusion model \cite{rombach2022high} for weight initialization, and trains a diffusion model specialized in instruction-guided editing. 
MagicBrush \cite{zhang2023magicbrush} fine-tunes IP2P using a real image dataset, thereby demonstrating a superior performance in instruction-guided editing.
In this paper, we aim to leverage the instruction-editing capability of these pretrained instruction-guided diffusion models to eliminate the need for additional masks in previous local editing approaches \cite{avrahami2022blended, avrahami2023blended, nichol2022glide}, enabling flexible and high-fidelity local editing based on a single user-provided instruction.

\begin{figure*}
    \centering
    \includegraphics[width=0.95\linewidth]{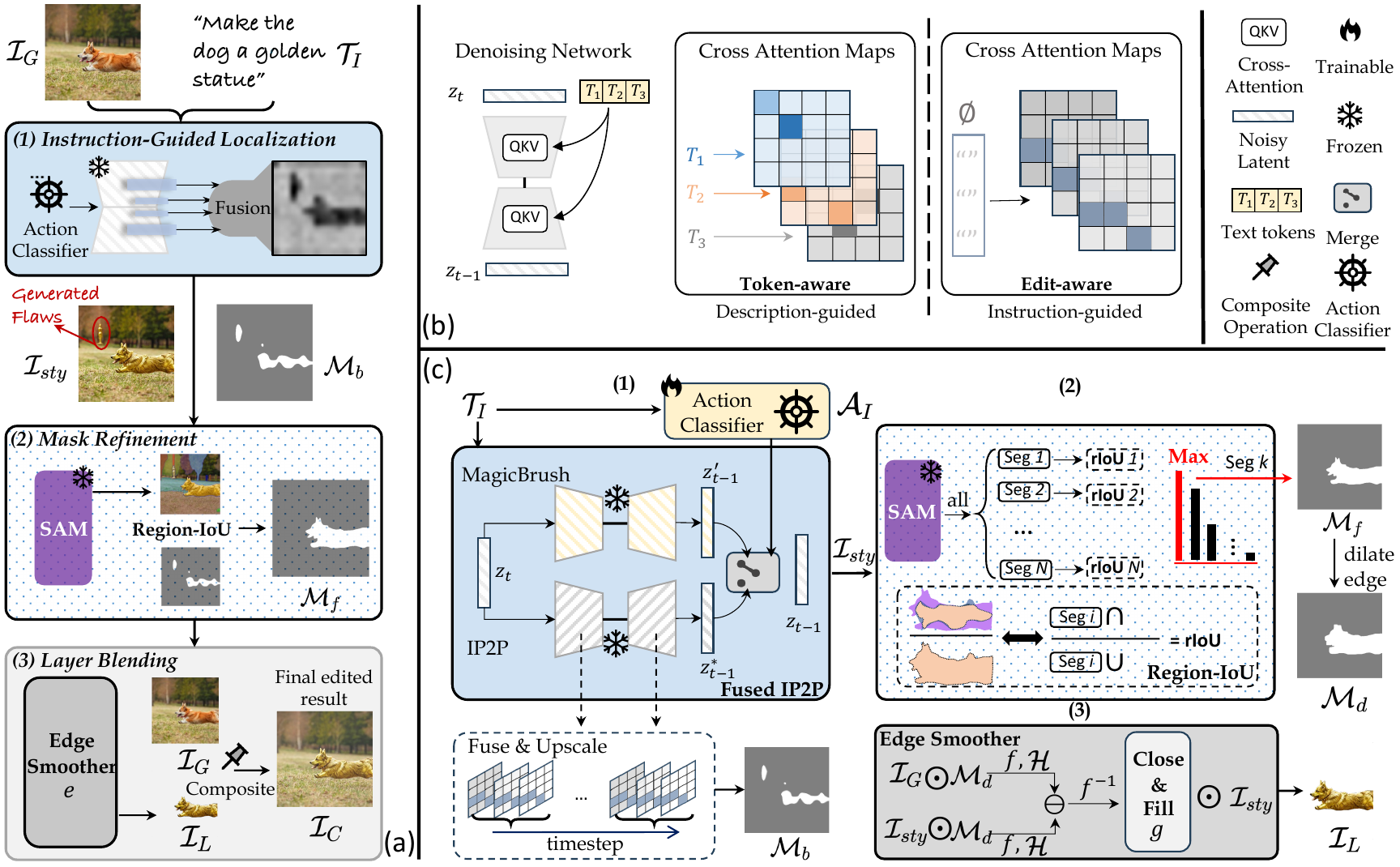}
    \caption{\textbf{Overview of \texttt{ZONE}}. (a) Three modules in \texttt{ZONE}. (b) The distinct difference between description-guided and instruction-guided diffusion models on cross-attention. The former usually follows a \textit{token-aware} format, while the latter is \textit{edit-aware}. $\varnothing$ denotes the unconditional embeddings for null input. (c) Implementation details of the modules shown in (a).}
    \label{fig:overview}
    \vspace{-3mm}
\end{figure*}
\section{Preliminaries}
\label{preliminary}
\paragraph{Diffusion Models.} 
Diffusion models \cite{sohl2015deep, ho2020denoising, song2020denoising} are probabilistic generative models founded on two complementary stochastic processes: \textit{diffusion} and \textit{denoising}. The \textit{diffusion} process progressively adds different amounts of Gaussian noise to a clean image $x_{0}$ towards Gaussian distribution $x_{T} \sim \mathcal{N}(0, I)$ in ${T}$ timesteps:
${x}_{t} = \sqrt{\alpha_t}{x}_{0}+ \sqrt{1-\alpha_t}\epsilon$,
where $\alpha_t$ defines the level of noise, and $\epsilon \sim \mathcal{N}(0, I)$. 

In the \textit{denoising} process, a neural network $\epsilon_{\theta}$ is designed to predict the noise $\epsilon$ for ${x}_{t}$ to get a ``cleaner" image gradually. This process is achieved by minimizing the denoising objective:
$\mathcal{L} = \mathbb{E}_{{x}_{0},t,\epsilon}\left \|\epsilon -\epsilon_{\theta}({x}_{t}, t)  \right \|^2_2$.
Rombach et al. \cite{rombach2022high} introduce a latent diffusion model (LDM), which speeds up both processes by reducing images into a lower-dimensional latent space utilizing a variational auto-encoder \cite{kingma2013auto}. This advancement has underpinned the achievements of Stable Diffusion, serving as the fundamental model for many diffusion-based works. 
\paragraph{InstructPix2Pix.} InstructPix2Pix \cite{brooks2023instructpix2pix} (IP2P) is a pioneering conditional diffusion model that edits images from user-provided instructions. Specifically, IP2P constructs an instruction dataset to fine-tune the pretrained Stable Diffusion. Given a target image $x$, an image condition $c_I$, and a textual instruction condition $c_T$, IP2P projects $x$ to the latent $z = \mathcal{E}(x)$ with a pretrained encoder $\mathcal{E}$, and then fine-tunes Stable Diffusion by minimizing the following objective: 
\begin{equation}
\begin{aligned}
\mathcal{L} & = \\
& \mathbb{E}_{\mathcal{E}(x), \mathcal{E}\left(c_{I}\right), c_{T}, \epsilon \sim \mathcal{N}(0,1), t}\left[\| \epsilon-\epsilon_{\theta}\left(z_{t}, t, \mathcal{E}\left(c_{I}\right), c_{T}\right)\right) \|_{2}^{2}],
\end{aligned}
\end{equation}
where the denoising network $\epsilon_{\theta}$ accepts two input conditions and predicts the noise $\epsilon$. IP2P also finds it beneficial to perform classifier-free guidance \cite{ho2022classifier} concerning both conditions, thus controlling the strength of edit by image guidance scale $s_I$ and instruction guidance scale $s_T$:
\begin{equation}
\begin{aligned}
\tilde{\epsilon_{\theta}}\left(z_{t}, c_{I}, c_{T}\right)= & \epsilon_{\theta}\left(z_{t}, \varnothing, \varnothing\right) \\
& +s_{I} \cdot\left(\epsilon_{\theta}\left(z_{t}, c_{I}, \varnothing\right)-\epsilon_{\theta}\left(z_{t}, \varnothing, \varnothing\right)\right) \\
& +s_{T} \cdot\left(\epsilon_{\theta}\left(z_{t}, c_{I}, c_{T}\right)-\epsilon_{\theta}\left(z_{t}, c_{I}, \varnothing\right)\right).
\end{aligned}
\end{equation}

At inference time, IP2P can modify an image with a user-provided instruction and trade-off the generated sample according to the strengths of the guidance image and the edit instruction through $s_I$ and $s_T$.


\section{Method}
\label{sec:method}

\paragraph{Overview of \texttt{ZONE}.}
We aim to make localized edits on an image with simple instructions. As depicted in Fig.~\ref{fig:teaser}~(a), such edits include performing three primary actions: (i) ``add": add an object to the image without specifying location with user-provided masks; (ii) ``remove": remove the object in the scene; (iii) ``change": change the style (\emph{i.e.,} texture) of an existing object or replace the object with another object. Additionally, our method allows high-fidelity multi-turn edits with a series of instructions. 

As outlined in Fig.~\ref{fig:overview}~(a), our approach consists of the following steps: First, we train an action classifier for steering different editing requirements and concurrently generate and position the editing region using a fused IP2P, as detailed in Section \ref{sec:locate} and Fig.~\ref{fig:overview}~(c). Second, we devise a mask refinement module for an edited image layer in Section \ref{sec:layer}. Finally, in Section \ref{sec:blend}, we propose an FFT-based edge smoother for seamless blending of the edited image layer with the original image.

\subsection{Problem Statement}
\label{sec:statement}
Given an RGB image $\mathcal{I}_{G} \in \mathbb{R}^{3 \times H \times W}$ and a textual instruction $\mathcal{T}_{I}$, we aim to locate and edit image regions following $\mathcal{T}_{I}$ and maintain the original non-edited regions. Inspired by Text2LIVE \cite{bar2022text2live}, we extract an edited layer $\mathcal{I}_{L}$ with color and opacity that are composited over $\mathcal{I}_{G}$. As opposed to previous works \cite{bar2022text2live, hertz2022prompt, avrahami2022blended, parmar2023zero}, we neither rely on any user-defined mask nor need non-intuitive prompt engineering, realizing precise local editing and seamless layer blending.

\subsection{Instruction-Guided Localization}
\label{sec:locate}
Many local editing methods require users to explicitly specify the object they want to edit with a prompt or a mask \cite{bar2022text2live, avrahami2022blended, couairon2022diffedit, parmar2023zero}. This is not intuitive and often requires a certain learning cost. Our approach locates and edits the implicitly designated object from the user's instruction. For example, a user-provided instruction like ``make her old" can implicitly convey the user's editing intent to modify the woman in the scene (\textit{locate}) by making her appear older (\textit{edit}).  

As shown in Fig.~\ref{fig:overview}~(b), our key finding is that 
the operational mechanisms of instruction-guided and description-guided diffusion models on cross-attention exhibit a distinct difference. 
Specifically, we empirically demonstrate that: (i) a description-guided model displays a \textit{token-aware} characteristic on its cross-attention maps, associating each input text token with a corresponding spatial structure; (ii) an instruction-guided model's cross-attention maps with unconditional embeddings share similar spatial features, demonstrating an \textit{edit-aware} characteristic, being responsive to the overall editing intent.

\begin{figure}[t]
    \centering
    \includegraphics[width=0.95\linewidth]{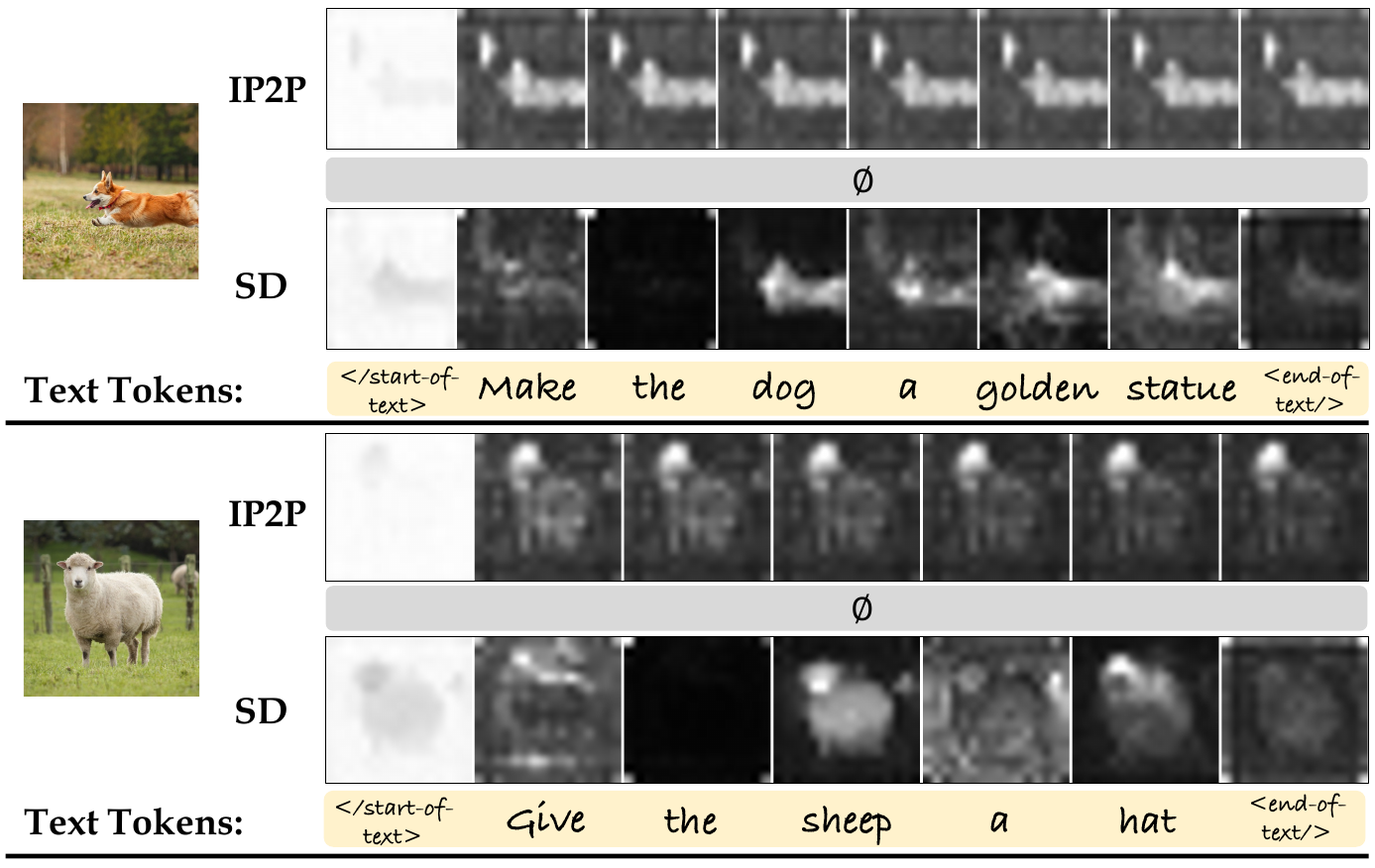}
    \caption{\textbf{Cross-attention map difference.} We average the cross-attention maps among all timesteps for each sample. IP2P shows consistency in the overall editing intent with unconditional embeddings $\varnothing$, while Stable Diffusion (SD) demonstrates a one-to-one correspondence with text tokens. 
    }
    \label{fig:diff}
    \vspace{-5mm}
\end{figure}

Given a noisy latent $z_t$ and a textual embedding $c_T$, the denoising UNet $\epsilon_\theta$ predicts the noise $\epsilon$ at each timestep $t$. 
The generation is conditioned on the textual prompt $\mathcal{T}_{I}$ by computing cross-attention between the textual embedding $c_T$ and the spatial features $\phi(z_t)$,  and updates $\phi(z_t)$ as $\hat{\phi}(z_t)$:
\begin{equation}
    M = \mathrm{Softmax}(\frac{QK^T}{\sqrt{d}}), \ \hat{\phi}(z_t) = M \cdot V,
\end{equation}
where the query $Q=W_{Q}\phi(z_t)$, key $K=W_{K}c_T$, and value $V=W_{V}c_T$ are obtained with linear projections $W_{Q}$, $W_{K}$, and $W_{V}$.
$M \in \mathbb{R}^{H^{'} \times W^{'} \times L}$ contains $L$ cross-attention maps that are correlated to the similarity between $Q$ and $K$. Typically $H^{'}$ and $W^{'}$ are $1/32$ of the original image size $H$ and $W$ in Stable Diffusion. 
For the description-guided Stable Diffusion model, each token corresponds to its specific attention map $M^{l}$ with text embeddings, where $l \in \{1,2,\ldots, L \}$. For the instruction-guided IP2P, we find its attention maps share a uniform characteristic with unconditional embeddings, concentrated directly at the edited location without token specification, as visualized in Fig.~\ref{fig:diff}.


Based on this finding, we devise a simple yet effective localization module that semantically locates the edited region with instruction $\mathcal{T}_{I}$. 
Specifically, we first collect the attention maps of the denoising model of IP2P from all timesteps of the denoising process.
Then, we average and resize the maps to obtain averaged attention maps $\mathcal{M}_{A} \in \mathbb{R}^{H \times W \times L}$. 
We observe that the first attention map $\mathcal{M}_{A}^{1}$ primarily emphasizes the attention weights of the non-edited region. Subsequent attention maps $\mathcal{M}_{A}^{2,..., L}$ shift attention towards the edited region. Therefore, we subtract the last cross-attention map from the first attention map and binarize each pixel $(m, n)$ with a fixed threshold $T$ to highlight the edited region and mitigate the background noise: 
\begin{equation}
    \mathcal{M}_{b}(m,n) = \begin{cases}
 1, \ \text{if } \mathcal{M}_{A}^{1}(m,n) - \mathcal{M}_{A}^{L}(m,n)<T, \\
0, \ \text{others, }
\end{cases}
\end{equation}
where $T$ is empirically set to $128$. This yields a rough, noise-filtered edited region mask $\mathcal{M}_{b}$ most related to $\mathcal{T}_{I}$ (see Fig.~\ref{fig:overview}~(a)).

Moreover, we find that IP2P performs not as well as MagicBrush in the ``remove" editing but preserves better object identity in terms of ``add" and ``change". 
Therefore, we design a fused IP2P module with a trainable action classifier $\mathcal{A}_{I}$. As illustrated in Fig.~\ref{fig:overview}~(c), we lock the weights of both IP2P and MagicBrush and use a pretrained action classifier $\mathcal{A}_{I}$ to steer the denoising process based on $\mathcal{T}_{I}$:
\begin{equation}
    z_{t-1} = (z_{t-1}^{*} + \beta \cdot z_{t-1}')/(1+\beta),
\end{equation}
where $z_{t-1}^{*}$ and $z_{t-1}'$ are the denoised latents by IP2P and MagicBrush, respectively. $\beta$ is a hyperparameter to control the guidance strength of MagicBrush on IP2P, empirically set to $0.2$ if $\mathcal{A}_{I}(\mathcal{T}_{I})$ is classified to ``remove" and $0.01$ for other actions. This module generates a globally edited image $\mathcal{I}_{sty}$ according to $\mathcal{T}_{I}$. $\mathcal{I}_{sty}$ serves as the canvas, from which the edited region is cropped out to form a separate image layer in the following steps.

\subsection{Mask Refinement}
\label{sec:layer}
The location mask $\mathcal{M}_{b}$ and $\mathcal{I}_{sty}$ obtained in Section~\ref{sec:locate} are insufficient for precise local editing,  since $\mathcal{M}_{b}$ only indicates the general location of the edited region, as illustrated in Fig.~\ref{fig:overview}~(a).
An intuitive and effective mask refinement method is to use an off-the-shelf segmentation model. 
We leverage the Segment Anything Model (SAM) \cite{kirillov2023segment} to generate precise masks of the canvas $\mathcal{I}_{sty}$ at various levels. 
However, we do not use SAM's preset point or box prompts for segmentation selection, because these prompts could potentially lead to misselection or omission of SAM's segmentation results due to IP2P's over-edit problem (which is also reflected in $\mathcal{M}_{b}$, see $\mathcal{I}_{sty}$ and $\mathcal{M}_{b}$ in Fig.~\ref{fig:overview}~(a)), resulting in a final mask that does not accurately reflect $\mathcal{T}_{I}$'s editing intention.
Therefore, we propose a Region-IoU (rIoU) scheme to obtain the accurate segmentation mask.

As depicted in Fig.~\ref{fig:overview}~(c), 
by sending $\mathcal{I}_{sty}$ to SAM, we extract all the possible instance segments $\mathcal{S} = \{\mathcal{S}^{j}\}^{N}_{j=1}$. Note that $\mathcal{S}$ contains the segments from all levels of SAM's segmentation. We define rIoU $\mathcal{R}(j)$ as:
\begin{equation}
    \mathcal{R}(j) = \frac{\mathrm{area}(\mathcal{S}^{j} \cap \mathcal{M}_{b})}{\mathrm{area}(\mathcal{S}^{j} \cup \mathcal{M}_{b})}, \ j=1,2,\ldots,N.
\end{equation}
If $k = \underset{j=1,2,\ldots, N}{\arg\max} \{ \mathcal{R}(j) \}$, then we obtain the refined mask $\mathcal{M}_{f}=\mathcal{S}^{k}$. One example is shown in Fig.~\ref{fig:overview}~(a) or (c).

\begin{table*}[htp]
\centering
\begin{tabular}{ccccccc}
\hline
\toprule
\textbf{Type}                               & \textbf{Methods} & L1 $\downarrow$ & L2 $\downarrow$ & LPIPS $\downarrow$ & CLIP-I $\uparrow$ & CLIP-T $\uparrow$ \\ 
\specialrule{0.5pt}{1pt}{1pt}
\multirow{3}{*}{\textbf{Description-guided}}      & DiffEdit \cite{couairon2022diffedit} & \underline{0.0426}  & 0.0099 &  \underline{0.1695} & 0.8947  & 0.2815  \\  
                                            & Text2LIVE \cite{bar2022text2live} & 0.0511 & \underline{0.0075} & 0.2176 & 0.9075 & \textbf{0.3062}  \\ 
                                            & Pix2Pix-Zero \cite{parmar2023zero} & 0.1198  & 0.0342  &  0.4375  & 0.7679 & 0.2701  \\ 
\specialrule{0.5pt}{1pt}{1pt}
\multirow{3}{*}{\textbf{Instruction-guided}} & InstructPix2Pix \cite{brooks2023instructpix2pix}  & 0.0945  &  0.0274  &  0.2816  &  \underline{0.9089} & 0.2907  \\ 
                                            & MagicBrush \cite{zhang2023magicbrush} & 0.0919  & 0.0378 & 0.2903  & 0.8959  &  0.2939  \\ 
                                            & \texttt{ZONE} (Ours)     &  \textbf{0.0146}  &  \textbf{0.0061} &  \textbf{0.0441} &  \textbf{0.9688}  &  \underline{0.2969} \\ 
\bottomrule
\end{tabular}
\caption{\textbf{Quantitative evaluation.} We use L1 and L2 to gauge pixel-level structural similarity, LPIPS and CLIP-I to evaluate image quality, and CLIP-T to assess text-image semantic similarity. The best and the second best results are marked in \textbf{bold} and \underline{underline}, respectively.}
\label{tab:quantitative}
\vspace{-3mm}
\end{table*}

\subsection{Layer Blending}
\label{sec:blend}

\begin{figure}[t]
    \centering
    \includegraphics[width=0.95\linewidth]{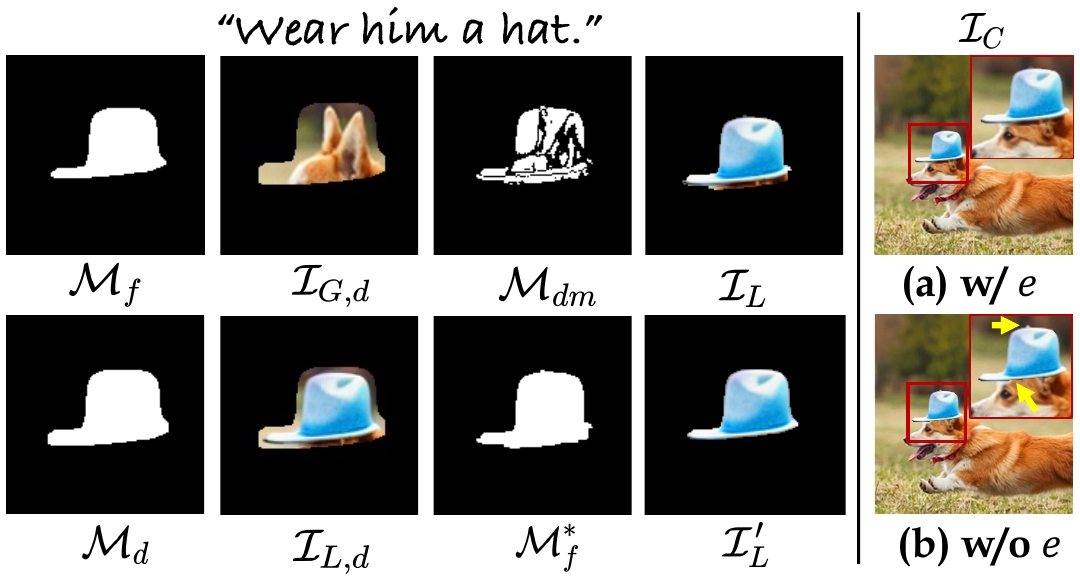}
    \caption{\textbf{Visualization and ablation.} The first 4 columns show the intermediate results related to the edge smoother. The last column compares the final edited results with and without the edge smoother.}
    \label{fig:edge}
    \vspace{-3mm}
\end{figure}

After the mask refinement, we obtain an edited image layer $\mathcal{I}_{L}' = \mathcal{I}_{sty} \odot \mathcal{M}_{f}$, which retains the color information of $\mathcal{I}_{sty}$ within the region where $\mathcal{M}_{f}=1$, with the rest being transparent. A naïve way to get the final edited result $\mathcal{I}_{C}$ is to stitch $\mathcal{I}_{L}'$ and the original image $\mathcal{I}_{G}$ at pixel-level. 
This fundamentally tackles the over-edit problem encountered in instruction-guided methods for local editing. Nevertheless, directly pasting $\mathcal{I}_{L}'$ back to $\mathcal{I}_{G}$ may result in noticeable artifacts, such as jagged edges and incomplete coverage of the edited region in the original image, as indicated by the yellow arrows in Fig.~\ref{fig:edge}~(b).


We tackle this problem by designing a novel edge smoother with Fast Fourier Transform (FFT). 
Given the original image $\mathcal{I}_{G}$, the canvas $\mathcal{I}_{sty}$, and the refined location mask $\mathcal{M}_{f}$, we first dilate $\mathcal{M}_{f}$ to $\mathcal{M}_{d}$ to incorporate more edge information in $\mathcal{I}_{sty}$ that may not be included in $\mathcal{I}_{L}'$. Then we get the dilated edited image layer $\mathcal{I}_{L,d} = \mathcal{I}_{sty} \odot \mathcal{M}_{d}$ and the dilated original image layer $\mathcal{I}_{G,d} = \mathcal{I}_{G} \odot \mathcal{M}_{d}$, as shown in the second column of Fig.~\ref{fig:edge}. The edge smoother $e$ is defined by:
\begin{equation}
    e(\mathcal{I}_{L,d}, \mathcal{I}_{G,d}) = g(f^{-1}(\mathcal{H}(f (\mathcal{I}_{L,d}))-\mathcal{H}(f (\mathcal{I}_{G,d})))), 
\end{equation}
where $g$ is a composition of binarization and morphological closing and filling functions, $f$ and $f^{-1}$ represent FFT and inverse FFT, respectively, and $\mathcal{H}$ is an ideal low-pass filter:
\begin{equation}
    \mathcal{H}(f_s) = \begin{cases} 
f_s(c), & \text{if } \| c - c_0 \|_2 \leq D_0, \\
0, & \text{if } \| c - c_0 \|_2 > D_0,
\end{cases}
\end{equation}
where $f_s \in \mathbb{R}^{H \times W}$ is the frequency spectrum of the image transformed by $f$, $c$ is the coordinate in $f_s$, $c_0$ is the center coordinate of $f_s$, and $D_0$ is set empirically to $200$ for a $512 \times 512$ image. We use the edge smoother $e$ to get the final mask $\mathcal{M}_{f}^{*}$.

As shown in the second column of Fig.~\ref{fig:edge}, 
we observe that both $\mathcal{I}_{G,d}$ and $\mathcal{I}_{L,d}$ share similar low-frequency characteristics on non-edited regions (\emph{e.g.,} background), but they hold different low-frequency characteristics on the edited regions (\emph{e.g.,} hat and the shadow below it). Therefore, we can exclude the non-edited regions and retain the edited regions by subtracting the low-frequency components between $\mathcal{I}_{L,d}$ and $\mathcal{I}_{G,d}$ in the frequency domain: $d_s=\mathcal{H}(f (\mathcal{I}_{L,d}))-\mathcal{H}(f (\mathcal{I}_{G,d}))$ and invert it back to the image domain to get the difference mask $\mathcal{M}_{dm}=f^{-1}(d_s)$. The final mask $\mathcal{M}_{f}^{*}$ is then obtained by $\mathcal{M}_{f}^{*}=g(\mathcal{M}_{dm})=e(\mathcal{I}_{L,d}, \mathcal{I}_{G,d})$. 
Finally, we get the final edited image layer $\mathcal{I}_{L}$ by $\mathcal{I}_{L} = \mathcal{I}_{sty} \odot \mathcal{M}_{f}^*$, and the final edited result $\mathcal{I}_{C}$ is acquired by compositing $\mathcal{I}_{G}$ and $\mathcal{I}_{L}$. The intuitive visualization of these intermediate results is shown in Fig.~\ref{fig:edge}.

The implementation details and more discussions can be found in the supplementary material.
\section{Experiments}
\label{sec:expri}

\begin{figure*}[htp]
    \centering
    \includegraphics[width=\linewidth]{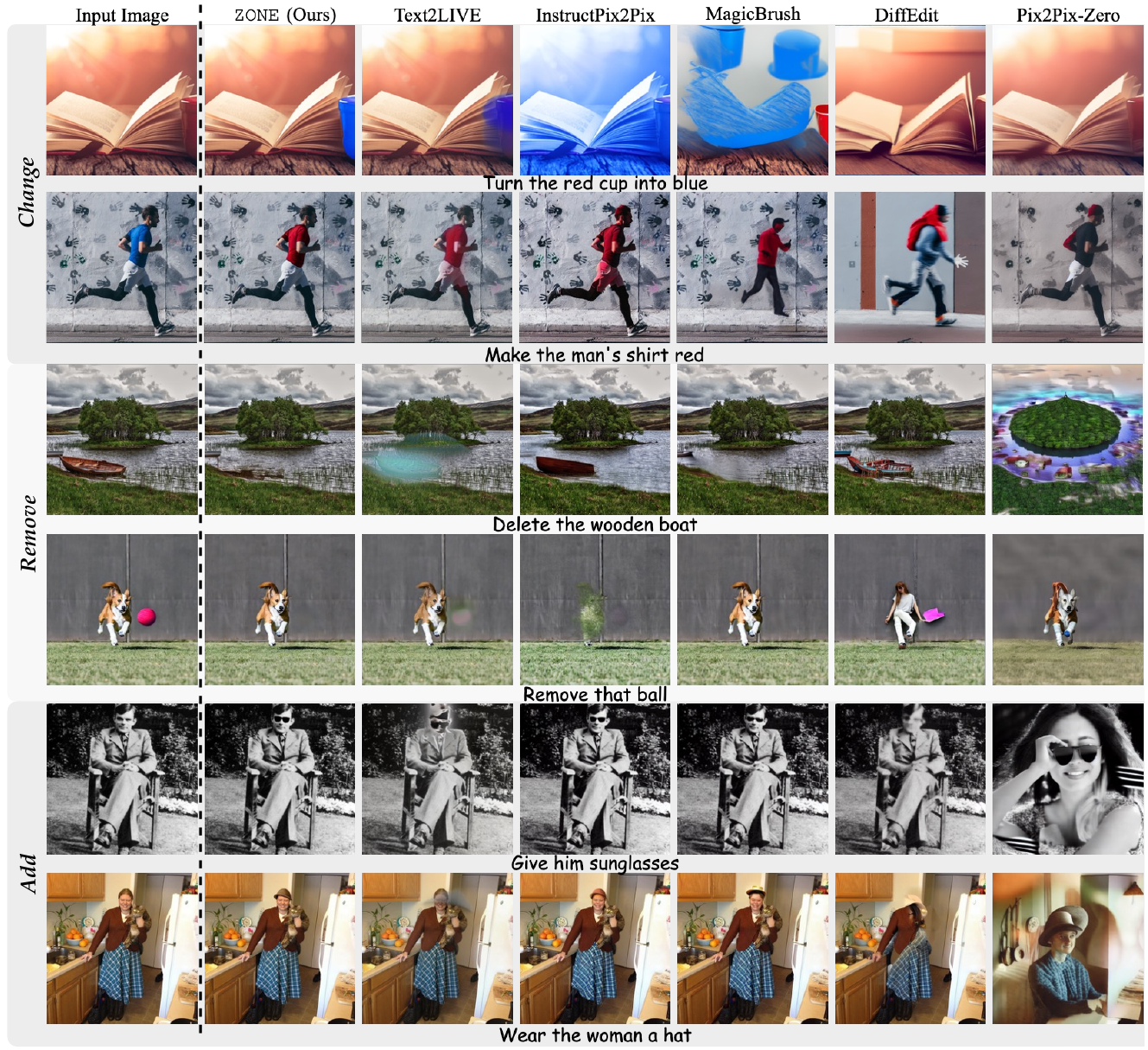}
    \caption{\textbf{Qualitative comparison.} We compare the editing efficacy of our \texttt{ZONE} with existing SOTA methods. The instructions (or instructions that are equivalent to the descriptions required by some baselines) used for editing are written below each row of the images. }
    \label{fig:qualitative}
    \vspace{-3mm}
\end{figure*}

\begin{figure}[htp]
    \centering
    \includegraphics[width=0.95\linewidth]{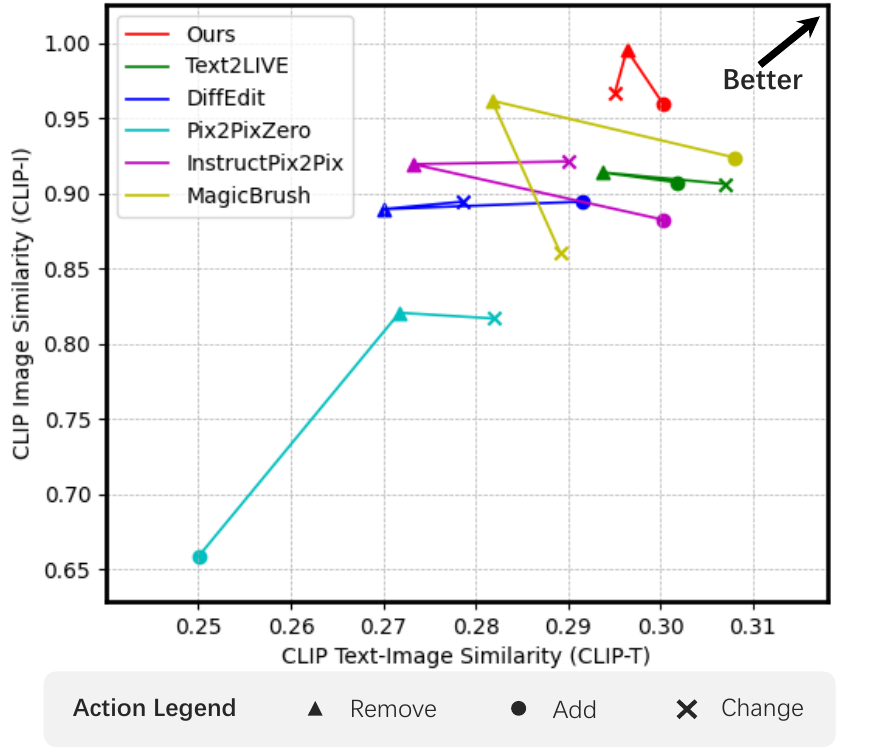}
    \caption{\textbf{Stability analysis.} We categorize the test set into three actions (``Remove", ``Add", and ``Change") and calculate their respective CLIP-I and CLIP-T values. Our method achieves the best quality-stability trade-off for all actions.}
    \label{fig:trade-off}
\end{figure}

\begin{figure}[htp]
    \centering
    \includegraphics[width=\linewidth]{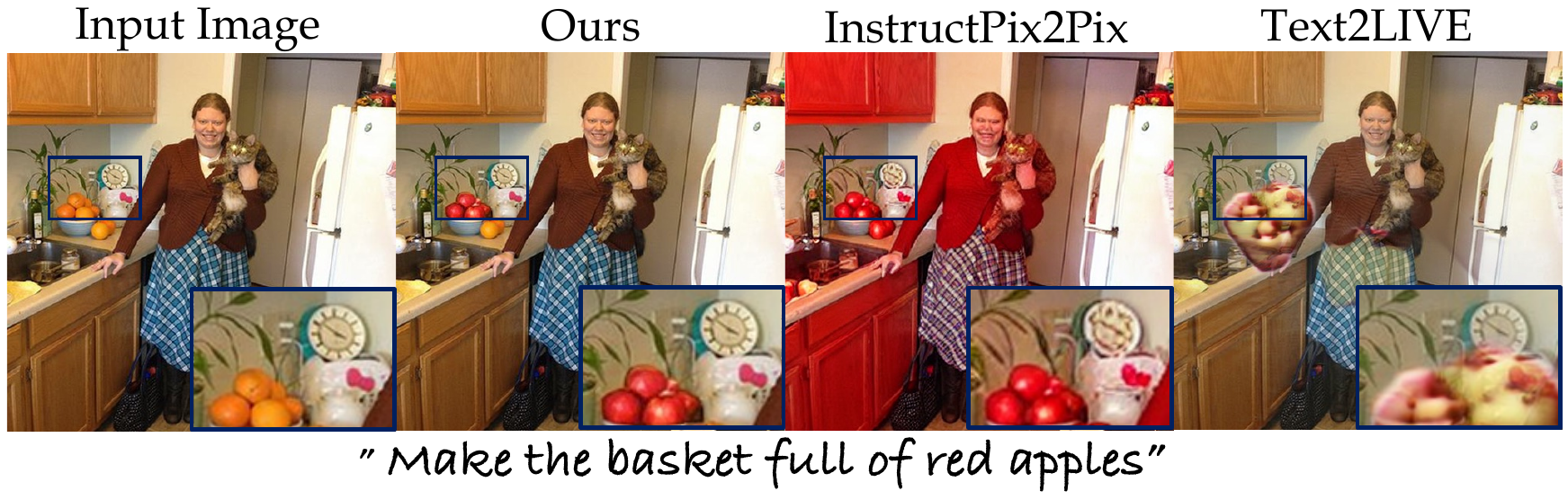}
    \caption{\textbf{Detailed comparison.} 
    We show a zoomed-in sample where \texttt{ZONE} effectively resolves the over-edit problem.}
    \label{fig:detail}
    \vspace{-3mm}
\end{figure}

\subsection{Experimental Setup}
\paragraph{Baselines.}
We conduct comprehensive experiments for the local editing task by comparing \texttt{ZONE} with five state-of-the-art image editing methods that are capable of local editing:
\textit{Text2LIVE} \cite{bar2022text2live},  \textit{DiffEdit} \cite{couairon2022diffedit}, \textit{IP2P} \cite{brooks2023instructpix2pix}, \textit{Pix2Pix-Zero} \cite{parmar2023zero}, and \textit{MagicBrush} \cite{zhang2023magicbrush}. The implementation of these methods can be found in the supplementary material.

\paragraph{Datasets.}
We randomly select and annotate 100 samples for evaluation, including 60 real images from the Internet and 40 synthetic images. To ensure the representativeness of the evaluation, we consider the diversity of scenes and objects in the sample selection. In particular, we divide the test set into three categories: 32 images for ``add", 54 for ``change", and 14 for ``remove" actions. All these 100 images are listed in the supplementary material.

\paragraph{Evaluation Metrics.}
\label{sec:metrics}
Following \cite{zhang2023magicbrush, brooks2023instructpix2pix}, we perform qualitative and quantitative comparisons using a variety of evaluation metrics. 
Learned Perceptual Image Patch Similarity (LPIPS) \cite{zhang2018unreasonable} is used to quantify the perceptual similarity between the original and edited image. CLIP text-image similarity (CLIP-T) \cite{gal2022stylegan} is employed to assess the alignment between the edited image and its corresponding caption, and CLIP image similarity (CLIP-I) is used to evaluate the layout similarity and semantic correlation between the edited image and the original image, serving as a reliable indicator of the edited image's quality. We also use L1 and L2 distances for pixel-level difference comparison.

\subsection{Comparisons}
\paragraph{Quantitative Evaluation.} As shown in Table~\ref{tab:quantitative}, we measure the models with the five metrics. 
The quantitative results indicate the following:
(i) Our method significantly outperforms our counterparts on metrics related to image structure and quality, implying the efficacy of \texttt{ZONE}'s preservation of the non-edited regions. 
(ii) Text2LIVE performs best on CLIP-T, but the qualitative comparison in Fig.~\ref{fig:qualitative} does not support this result. 
We surmise that Text2LIVE performs better on this metric potentially due to its direct supervision by CLIP. 

To quantify the stability of the edits, we divide the test set into three action groups: ``change", ``add", and ``remove". We then test the CLIP-I and CLIP-T metrics for each model and plot the CLIP curves in Fig.~\ref{fig:trade-off}, where the performances of the same method on these actions are connected with lines of the same color. Our interpretation is as follows: first, the shorter the projection of the line on the axis, the higher the semantic \textit{stability} (\emph{i.e.,} maintaining similar performances under different editing instructions) of the image editing; second, if the curve is closer to the upper right corner, it indicates that the method's editing \textit{quality} is more superior.
Our method achieves the best trade-off between \textit{quality} and \textit{stability}, 
demonstrating strong editing stability and representativeness.
\vspace{-3mm}

\paragraph{Qualitative Comparsion.} In Fig.~\ref{fig:qualitative}, we illustrate the editing results for the baselines and our method. We select six sets of images (including synthetic and real images) and group them based on actions. Our \texttt{ZONE} shows precise local editing capability while preserving the remaining pixels, this is especially important when there are perceptually important high-frequency details, such as faces, textures, or texts. 
A zoomed-in comparison is shown in Fig.~\ref{fig:detail}.
Both InstructPix2Pix and Text2LIVE introduce distortions to the non-edited areas during the editing process. For instance, InstructPix2Pix distorts the nearby clock and paints the orange outside of the basket red. In comparison, Text2LIVE maintains a better structure but generates a ``barrel" of apples and introduces an obvious foggy effect to the image.
Our method, however, can clearly distinguish between the edited region and the non-edited regions, demonstrating the best local editing efficacy.

\begin{table}[]
\centering
\begin{tabular}{ccc}
\toprule
\multirow{1}{*}{\textbf{Methods}} &  
                          SR (\%) & UPR (\%) \\ 
\specialrule{0.5pt}{1pt}{1pt}
DiffEdit  \cite{couairon2022diffedit}  &  27.1 $\pm$ 2.7  &  8.8 \\
Text2LIVE  \cite{bar2022text2live}     &  33.0 $\pm$ 3.2  &  17.3 \\
Pix2Pix-Zero \cite{parmar2023zero}      &  19.2 $\pm$ 3.7  & 10.4 \\
InstructPix2Pix  \cite{brooks2023instructpix2pix}  &   59.8 $\pm$ 3.1   & 18.9   \\
MagicBrush  \cite{zhang2023magicbrush}    &   50.2 $\pm$ 2.9   & 18.0  \\
\texttt{ZONE} (Ours) &   \textbf{69.4} $\pm$ \textbf{3.5}  & \textbf{26.6}\\ 
\bottomrule
\end{tabular}
\caption{\textbf{Human evaluation.} Our \texttt{ZONE} obtains the highest success rate (SR) and user preference rate (UPR).}
\label{tab:success}
\vspace{-4mm}
\end{table}





\subsection{Human Evaluation}
Due to the lack of an effective metric to measure editing effects (mainly due to the absence of ground truth images after editing), the metrics mentioned in Section~\ref{sec:metrics} alone are not sufficient to demonstrate the superiority of our method over existing ones. To further validate the editing effects of \texttt{ZONE}, in addition to the visual comparison in Fig.~\ref{fig:qualitative}, we also conduct a human evaluation to calculate the success rate (SR) and user preference rate (UPR) of the edited images with the editing instructions. Table~\ref{tab:success} shows a consistent preference for our method by users, as well as a dominant success rate over other methods. 

Please refer to our supplementary material for more visualizations and details of this user study.


\section{Conclusion}
\label{sec:concl}
We present \texttt{ZONE}, a zero-shot instruction-guided local image editing approach, which leverages the localization capability within the pre-trained instruction-guided diffusion models. Our approach innovatively utilizes the editing intent regions inherent in the instructions, rather than focusing on individual tokens, eliminating the need for specific guidance. By integrating the Region-IoU scheme and FFT-based edge smoother with a pretrained segmentation model, \texttt{ZONE} effectively realizes precise local editing. Comprehensive experiments and user studies further demonstrate the superiority of \texttt{ZONE} over SOTA methods. \\

\textbf{Acknowledgements.}
The work was supported by the National Key Research and Development Program of China (2023YFC3300029), Zhejiang Provincial Natural Science Foundation of China (LD24F020007), Beijing Natural Science Foundation (L223024), National Natural Science Foundation of China (62076016), “One Thousand Plan” projects in Jiangxi Province (Jxsg2023102268), Beijing Municipal Science \& Technology Commission, Administrative Commission of Zhongguancun Science Park (Z231100005923035).

{
    \small
    \bibliographystyle{ieeenat_fullname}
    \bibliography{main}
}
\newpage
\onecolumn

\setcounter{section}{0}
\renewcommand{\thesection}{\Alph{section}}

\section*{\LARGE\textbf Appendix}

In this supplementary, we first give more visualization results, then detail the datasets and the implementation, and finally discuss the limitations and state the social impact.

\section{More Visualizations}
In this section, we first present more visualizations of the samples from the test set under two comparison settings: (i) single-turn editing, and (ii) multi-turn editing. To make the comparison more representative, we compare our \texttt{ZONE} with three state-of-the-art (SOTA) text-to-image approaches, Text2LIVE (T2L) \cite{bar2022text2live}, InstructPix2Pix (IP2P) \cite{brooks2023instructpix2pix}, and MagicBrush (MB) \cite{zhang2023magicbrush}. 
Then we conduct an ablation study to show the efficacy of our fused IP2P module, through cross-attention map visualization. 

\subsection{Single-Turn Editing Examples}

\begin{figure}[htp]
    \centering
    \includegraphics[width=0.87\linewidth]{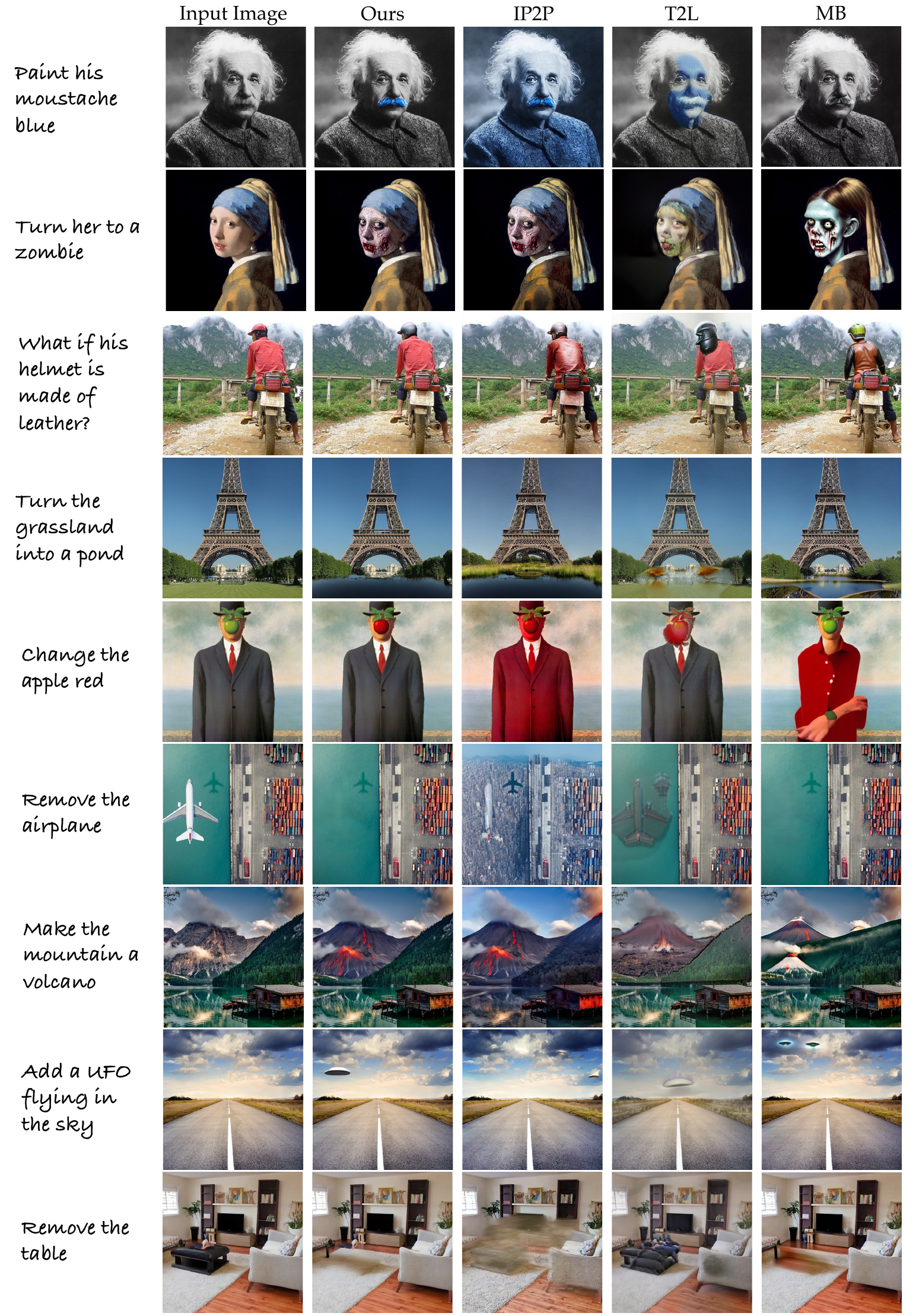}
    \caption{Single-turn editing examples. IP2P: InstructPix2Pix \cite{brooks2023instructpix2pix}; T2L: Text2LIVE \cite{bar2022text2live}; MB: MagicBrush \cite{zhang2023magicbrush}.}
    \label{fig:single-turn}
\end{figure}

We show more single-turn editing examples to further validate \texttt{ZONE}'s remarkable ability of local image editing. In particular, we compare it with the other methods for local editing using 9 images and their corresponding instructions (or prompts equivalent to the instructions). As evident in Fig.~\ref{fig:single-turn}, the results generated by \texttt{ZONE} surpass those of the other methods, demonstrating its impressive prowess in local editing.

\subsection{Multi-Turn Editing Examples}

\begin{figure}
    \centering
    \includegraphics[width=\linewidth]{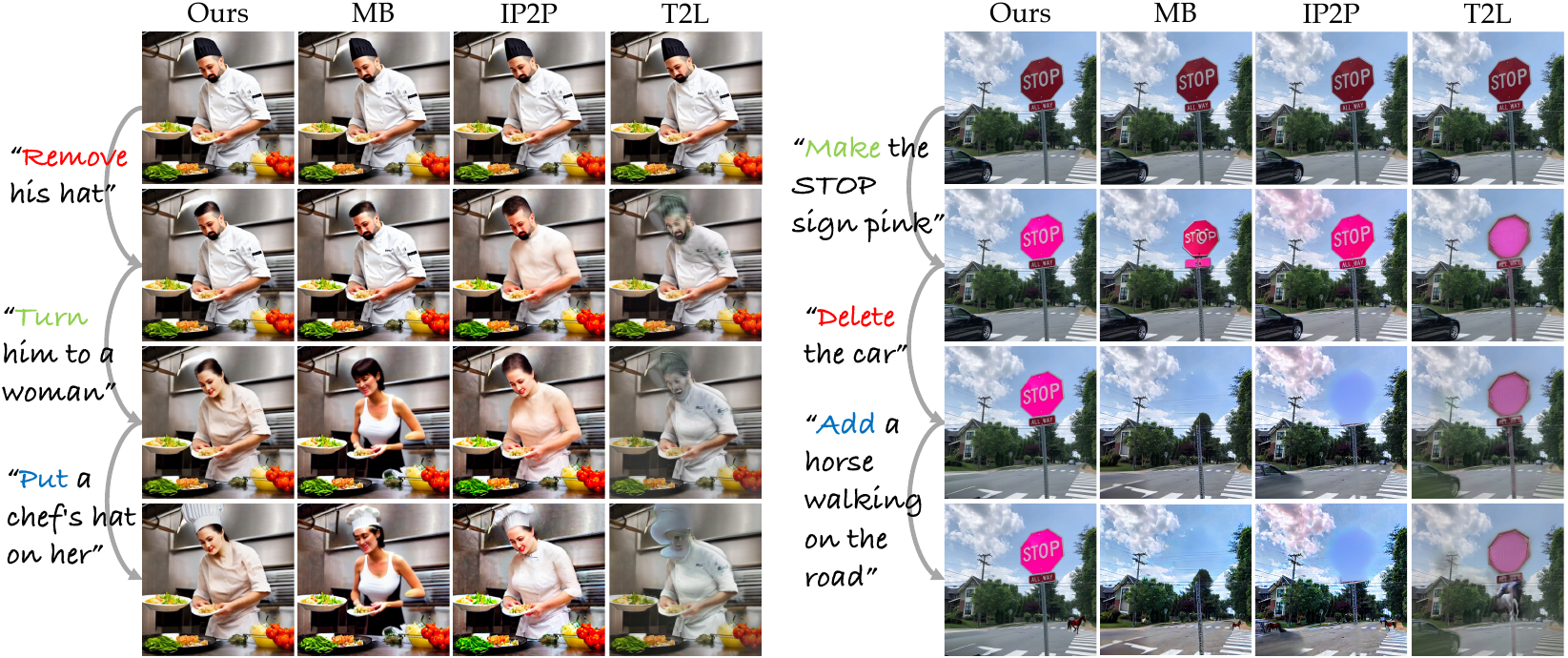}
    \caption{Multi-turn editing examples. IP2P: InstructPix2Pix \cite{brooks2023instructpix2pix}; T2L: Text2LIVE \cite{bar2022text2live}; MB: MagicBrush \cite{zhang2023magicbrush}. Best viewed zoomed in.}
    \label{fig:multi-turn}
\end{figure}

We use our \texttt{ZONE} to edit 2 images in a multi-turn style and compare the editing results with those obtained from the other methods. Specifically, each method is employed to edit each image three times, with different instructions. As illustrated in Fig.~\ref{fig:multi-turn}, our \texttt{ZONE} can achieve high-quality local edits under multiple instructions and preserve the original image's non-edited regions. In contrast, the results generated by the other methods exhibit noticeable distortions from the original images after multiple rounds of editing, which is not preferred in practical applications.

\subsection{Cross-Attention Map Visualization}

\begin{figure}[h]
    \centering
    \includegraphics[width=\linewidth]{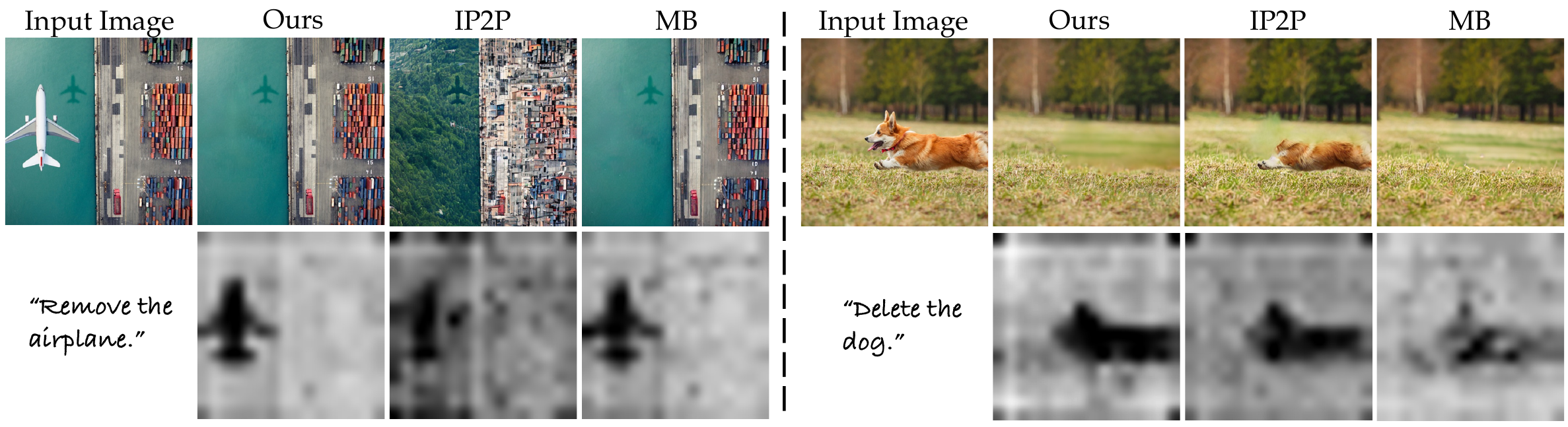}
    \caption{Cross-attention map comparisons. The \textbf{darker parts} in each cross-attention map (the second row) denote the edit regions.}
    \label{fig:crossvis}
\end{figure}

As shown in Fig.~\ref{fig:crossvis}, the first row demonstrates the editing results, and the second row illustrates the averaged cross-attention maps. From the cross-attention maps, we can see that by fusing the denoised latents of the two methods as described in Equation (5) of the main paper, our approach achieves a better localization capability under the ``Remove" editing intent compared to the two methods. 

\section{Implementation Details}

We conduct all our experiments based on open-source projects and models. We adopt an NVIDIA V100-SXM2-32GB GPU for the action classifier training and for $\texttt{ZONE}$ testing. 
The action classifier $\mathcal{A}_{I}$ leverages the instruction embeddings extracted by the text encoder of InstructPix2Pix (IP2P) \cite{brooks2023instructpix2pix} as its input, and outputs the probability logits for each action. To train the action classifier, we first use GPT-3.5 to generate samples for training and testing, and then we lock the weights of the text encoder of IP2P and optimize $\mathcal{A}_{I}$ using Adam \cite{kingma2014adam} with a learning rate of 0.1 for 30 epochs. The action classifier achieves 100\% top-1 classification accuracy on the test set.

We set 20 sampling steps for the fused IP2P and average the cross-attention layers of the first three UNet upsampling blocks and the second to the fourth downsampling blocks to get the fused cross-attention maps among all the denoising steps.

The action classifier $\mathcal{A}_{I}$ is a simple Multi-Layer Perceptron (MLP), comprising two linear layers with an intermediate ReLU activation function. The input dimension of the first linear layer and the length of the embedding outputted from the CLIP text encoder \cite{radford2021learning} are the same (equal to 768), and the output dimension at this layer is 128. The intermediate ReLU function introduces non-linearity to the output, and the second linear layer takes the 128-dimensional output from the ReLU function and produces a 3-dimensional output to classify the given instruction. 


\section{Experimental Details}
\subsection{Baselines}
To ensure consistency and convenience in method comparison, we uniformly adopt the implementation from the diffusers project \footnote{https://github.com/huggingface/diffusers} for IP2P \cite{brooks2023instructpix2pix}, MagicBrush \cite{zhang2023magicbrush}, DiffEdit \cite{couairon2022diffedit}, and Pix2Pix-Zero \cite{parmar2023zero}, and use their default parameters to generate results and calculate the metrics.
For Text2LIVE \cite{bar2022text2live}, we conduct experiments using its official code repository. To eliminate the potential discrepancies in generative capabilities arising from different versions of Stable Diffusion used across these methods, we employ Stable Diffusion 1.5 \footnote{https://huggingface.co/runwayml/stable-diffusion-v1-5} as the base model. Notably, since half of these methods do not support instructions as textual inputs, we design text prompts or additional assistance equivalent to instructions during our comparative experiments to achieve a relatively fair comparison.

\subsection{Datasets}
In this section, we provide the generation details of the dataset for action classification and the test set that we collect to evaluate the metrics for our \texttt{ZONE} and other editing methods.

\paragraph{Dataset for action classification.}
We employ GPT-3.5 \footnote{https://chat.openai.com} to generate the dataset used for training the action classifier. Our primary objective is to generate sentences that closely resemble user instructions, with the editing focus on common items found in real images. To achieve this, we choose categories from the COCO dataset to serve as the vocabulary for sentence generation. The following prompt is designed for generating training and testing data:
\\

\textsf{``Now you are a dataset bot, who will generate a training dataset for a three-fold (change, add, and remove) sentence classification task. 
    Specifically, you should generate a sentence along with its label. 
    In this task, we aim to generate a dataset for ``change", ``add", and ``remove" (labeled 0, 1, 2): 
    For example: ``turn the cat into a dog, 0", ``give the dog a hat, 1", ``get rid of the person on the left, 2".
    You should generate 450 sentence-label pairs if I give the instruction ``train", and 150 pairs when I give the instruction ``test". 
    I expect your response to be straight-forward, each sentence should be within 30 words, and you should freely select the words in the following list: 
    [
    `person', `bicycle', `car', `motorcycle', `airplane', `bus', `train', `truck', `boat', `traffic light',
    `fire hydrant', `stop sign', `parking meter', `bench', `bird', `cat', `dog', `horse', `sheep', `cow',
    `elephant', `bear', `zebra', `giraffe', `backpack', `umbrella', `handbag', `tie', `suitcase', `frisbee',
    `skis', `snowboard', `sports ball', `kite', `baseball bat', `baseball glove', `skateboard', `surfboard',
    `tennis racket', `bottle', `wine glass', `cup', `fork', `knife', `spoon', `bowl', `banana', `apple',
    `sandwich', `orange', `broccoli', `carrot', `hot dog', `pizza', `donut', `cake', `chair', `couch',
    `potted plant', `bed', `dining table', `toilet', `tv', `laptop', `mouse', `remote', `keyboard', `cell phone',
    `microwave', `oven', `toaster', `sink', `refrigerator', `book', `clock', `vase', `scissors', `teddy bear',
    `hair drier', `toothbrush'
    ]
    and make sure the sentence is short and clear, and the label is correct, and the dataset is balanced.
    Please just reply with the sentence-label pairs, and wait for my instructions. Note that the generated sentence-label pairs should not repeat."}
\\

The data generated by GPT-3.5 undergoes manual verification. The final training dataset includes 150 samples each for the ``add", ``remove", and ``change" actions, while the test dataset comprises 50 samples for each action. We show some samples from the training dataset in Table~\ref{tab:example_dataset}.

\begin{table}[htp]
\centering
\begin{tabular}{c}
\toprule
Turn the bicycle into a motorcycle, 0          \\
Make the apple a banana, 0                     \\
Swap the baseball glove for a tennis racket, 0 \\
Replace the chair with a couch, 0              \\ \hline
Put a frisbee next to the cat, 1               \\
Attach a remote to the TV, 1                   \\
Include a toothbrush on the dining table, 1    \\
Give the horse a suitcase, 1                   \\ \hline
Remove the horse, 2                            \\
Take away the umbrella, 2                      \\
Delete the traffic light, 2                    \\
Erase the microwave, 2                         \\ 
\bottomrule
\end{tabular}
\caption{Examples of the training dataset of the action classifier.}
\label{tab:example_dataset}
\end{table}

\begin{figure}[h]
    \centering
    \includegraphics[width=\linewidth]{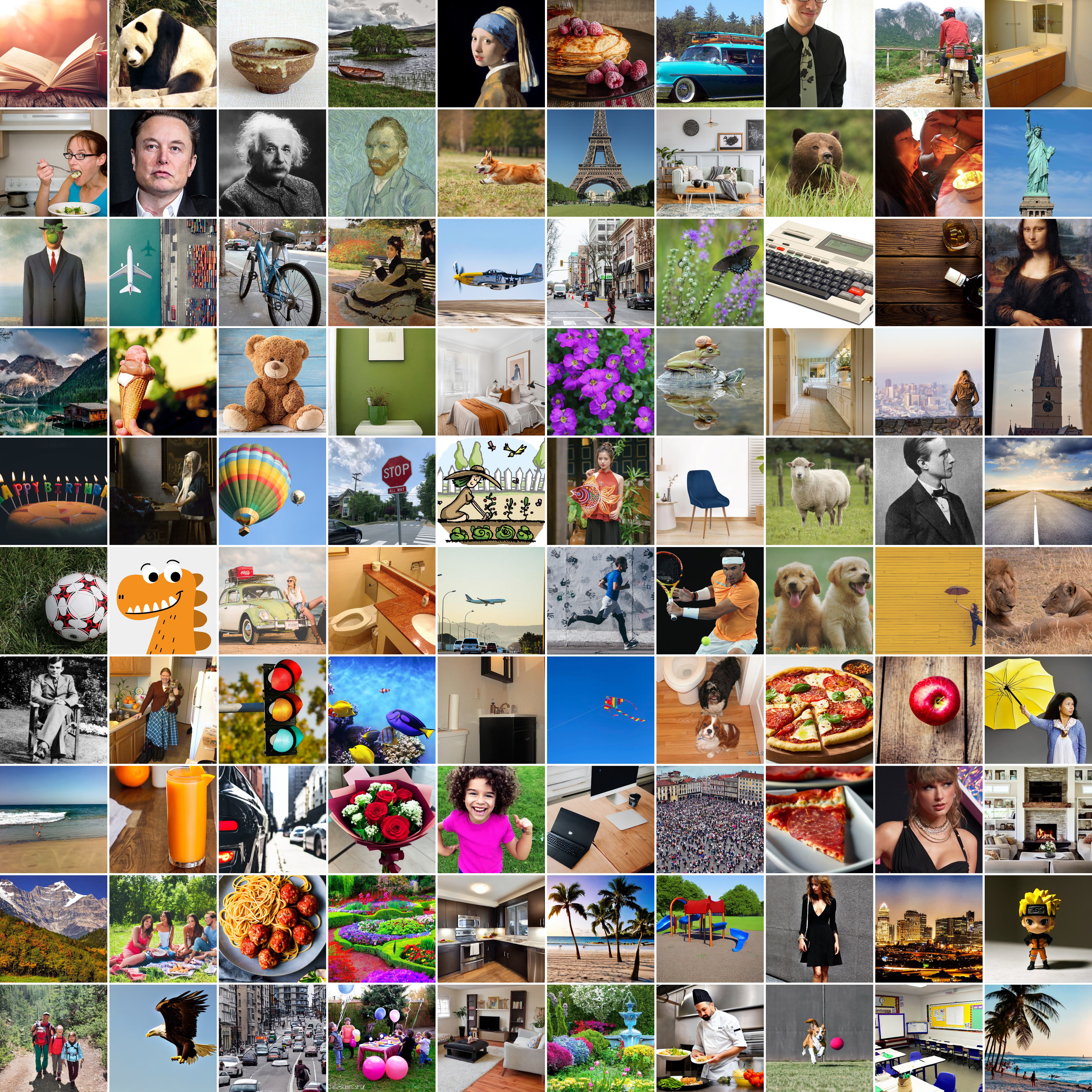}
    \caption{\textbf{Images in the test set}. We calculate the evaluation metrics and provide visualizations in the main paper with images in this set.}
    \label{fig:testset}
\end{figure}

\paragraph{Test set for evaluation.} 
We present the test set utilized in our evaluation in Fig.~\ref{fig:testset}. Initially, we gather 60 images from the Internet and create 40 synthetic images using Stable Diffusion 1.5 \cite{rombach2022high}. Subsequently, each image is cropped to a resolution of $512 \times 512$. Then we use BLIP \cite{li2022blip} to caption each image and manually annotate the instructions, output captions, source objects, and target objects. Three annotation examples are shown in Table~\ref{tab:anno}.

\begin{table}[htp]
\centering
\begin{tabular}{c|ccc}
\toprule
\textbf{Keys}   & \textbf{Example 1}                 & \textbf{Example 2}                         & \textbf{Example 3}                                   \\ \hline
action          & Change                          & Remove                                  & Add                                               \\
input caption  & A blue car in front of a forest & A man in black with a tie & A photo of Elon Musk                      \\
output caption & A red car in front of a forest  & A man in black                         & A photo of Elon Musk with glasses \\
instruction     & paint the car red               & get off his tie                         & give him glasses                          \\
source object  & blue car                        & a tie                             & N/A                                               \\
target object  & red car                         & N/A                                     & glasses                                             \\
\bottomrule

\end{tabular}
\caption{Three annotation examples. ``N/A" indicates the absence of words.}
\label{tab:anno}
\end{table}


\subsection{Evaluation Metrics}
\paragraph{L1/L2 distance.} The L1 and L2 distances serve as the metrics for evaluating structural and pixel-wise similarities between two images. The L1 distance measures absolute differences in pixel values, while the L2 distance calculates squared differences. Both metrics play a critical role in assessing dissimilarity, with smaller distances indicating greater image similarity in both pixel intensity and spatial structure.

\paragraph{LPIPS score.} LPIPS (Learned Perceptual Image Patch Similarity) \cite{zhang2018unreasonable} is a metric designed for evaluating the perceptual similarity between two images. It takes into account both pixel-level differences and high-level visual features, providing a comprehensive measure of how similar images appear to humans. 

\paragraph{CLIP-based metrics.} 
CLIP, or Contrastive Language-Image Pre-training, is a transformative model that excels in understanding the intricate relationships between text descriptions and images \cite{radford2021learning}. Through a pre-training process that employs contrastive learning, CLIP learns a shared embedding space where images and text descriptions are represented as vectors. This shared space is designed to bring semantically related content in close proximity. The model tokenizes images into regions and text into tokens, leveraging a transformer architecture with cross-modal attention to establish connections between corresponding regions and tokens. 
Both the CLIP-I and CLIP-T metrics evaluate the input image/text in the shared embedding space:

\begin{itemize}
    \item CLIP image similarity (CLIP-I) is designed to evaluate the image quality in both semantics and structure. This metric is computed by calculating the cosine similarity of the embedding vectors of the source image and the target image. 

    \item CLIP text-image similarity (CLIP-T) is used to evaluate the alignment between the edited image and its corresponding caption. More specifically, CLIP-T calculates the cosine similarity between the embedding vectors of the edited image and its corresponding caption.
\end{itemize}

\paragraph{More evaluation details.}We employ $512 \times 512$ images as inputs for each method during evaluations. However, DiffEdit \cite{couairon2022diffedit} requires image inputs with a resolution of $768 \times 768$ to function properly. So we first resize the test images to $768 \times 768$ for DiffEdit to ensure its proper performance and resize the outputs back to $512 \times 512$ to calculate the metrics.


\subsection{Human Evaluation}
\begin{figure}[h]
    \centering
    \includegraphics[width=0.95\linewidth]{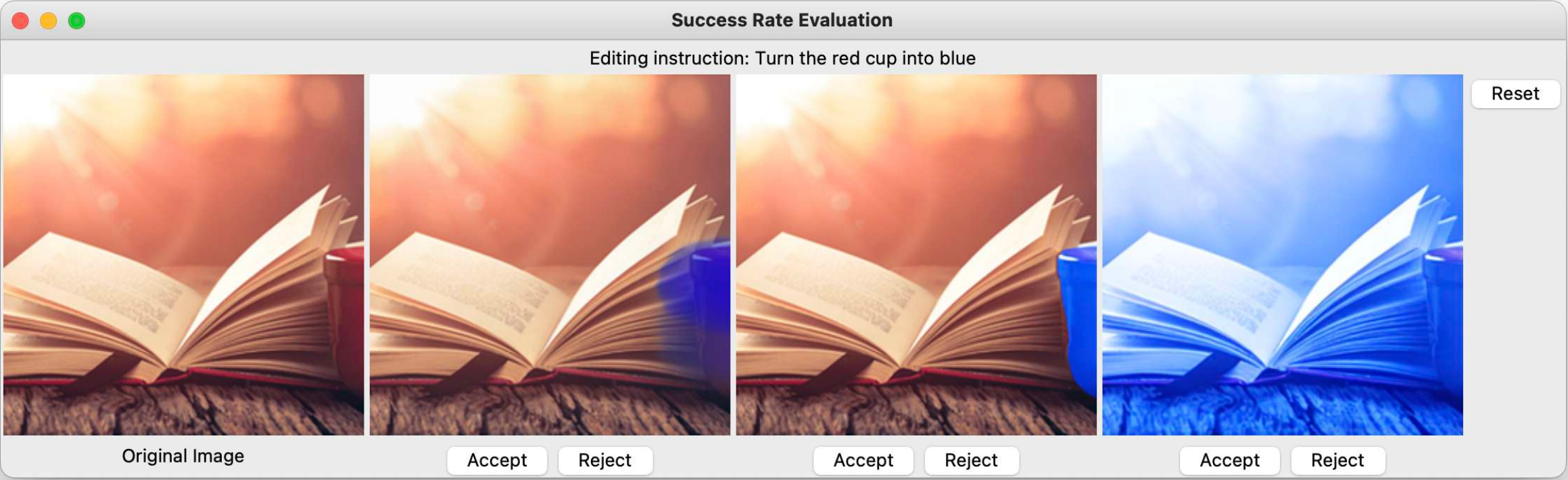}
    \vspace{-3mm}
    \caption{A screenshot of the annotation tool. The original image, the instruction, and three results randomly selected from the six methods are displayed each time.}
    \label{fig:software}
\end{figure}

\paragraph{Success rate.} We invite five volunteers to annotate the success rates of the six methods on the test set. To simplify the annotation process and avoid bias, we design a tool that can display the editing results of each method in a randomly shuffled order and anonymous style (see Fig.~\ref{fig:software}). The volunteers are then asked to decide whether to accept or reject the edited image based on the editing quality (\emph{i.e.,} preservation of the non-edited regions and the realism of the edited image) and text-image alignment between the output caption and the edited image. Ultimately, the success rate of each method is obtained by dividing the number of accepted results by the total number. To minimize annotation bias, we calculate the mean and standard deviation of the success rates from the five volunteers and demonstrate the results in Table 2 of the main paper. 

\paragraph{User preference rate.} We conduct a user study, which includes 16 sets of randomly selected editing results. Each set contains six results obtained by the six methods that we compare in the experiment, presented in a randomly shuffled order. The users are asked to give a preference score according to the degree of agreement between each editing result and the corresponding instruction, as well as the similarity to the original image, with the score from 1 to 10 and a higher score indicating a higher preference. A total of 30 users participate in this test. The final results are calculated by dividing the total score obtained for each method $S_{i}$ by the total score obtained for all methods:
\begin{equation}
    UPR(i) = 100 \times S_{i}/\sum_{i=1}^{6}{S_{i}}.
\end{equation}

\subsection{Additional Ablations}
We perform further ablations to demonstrate the effectiveness of each component within our proposed \texttt{ZONE}. First, we ablate on the $\beta$ parameter of the fused IP2P module. $\beta$ is a hyperparameter that controls the guidance strength of MagicBrush (MB) on the fused IP2P module. A higher $\beta$ emphasizes MB's effects on the editing results (please check L302-314 in our paper for more details). We illustrate the results generated by the models under various $\beta$ in Fig.~\ref{fig:betaparam}:

\begin{figure}[h]
    \centering
    \includegraphics[width=0.92\linewidth]{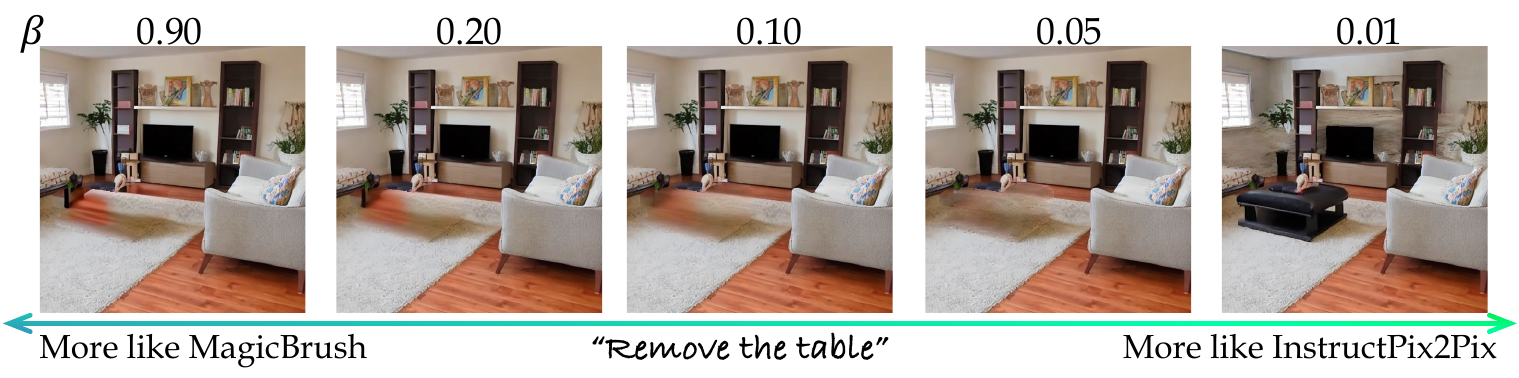}
    \caption{\textbf{Ablation on $\beta$.} As $\beta$ decreases from left to right, the generated result increasingly resembles that produced by InstructPix2Pix and less resembles that generated by MagicBrush. Given the limits of current instruction-guided diffusion models, we employ this character to handle different editing actions.}
    \label{fig:betaparam}
\end{figure}

Then we conduct an ablation study on the components of \texttt{ZONE}, including SAM and the edge smoother, with human evaluation using the user preference rate (UPR) and the CLIP image similarity (CLIP-I) for metrics evaluation. As demonstrated in Tab.~\ref{tab:component_ablat}, the implementation of SAM notably boosts the user preference rate, signifying a visually discernible enhancement compared to its absence, a fact also mirrored in the CLIP-I score. Concurrently, the user preference rate exhibits a significant improvement with the integration of the Edge Smoother, underscoring its efficacy.
\begin{table}
\centering
\begin{tabular}{c|cc|cc}
\hline
\toprule
        \multirow{2}{*}{Component}  & \multicolumn{2}{c}{SAM} & \multicolumn{2}{c}{Edge Smoother} \\ 
                                   & w/o                & w/             & w/o              & w/             \\ \midrule
        CLIP-I $\uparrow$                     & 0.95           & \textbf{0.97}         & \textbf{0.97}                & \textbf{0.97}              \\
        UPR (\%) $\uparrow$              & 5.4           & \textbf{94.6}        & 22.7               & \textbf{77.3}           \\
        \bottomrule
\end{tabular}
\caption{\textbf{Ablation on \texttt{ZONE}'s components.} We evaluate both components with human evaluation and CLIP metrics. }
\label{tab:component_ablat}
\end{table}


\section{Limitations}
While our method can produce impressive local manipulations of images and address the over-edit issue of InstructPix2Pix, it still has limitations. 
First, its editing capabilities are constrained by the instruction-guided diffusion models we employ, which may lead to occasional ineffectiveness in editing. This issue can be addressed in the future with more powerful instruction-guided diffusion models.
Secondly, our method falls short of localization in complex scenes (\emph{e.g.,} multiple similar objects or tiny objects), which is a challenging task that still needs to be explored.
Lastly, the current set of editing actions is relatively limited, more actions like ``move", ``resize", or ``copy" will be considered in future work. 

\section{Social Impact}
Our work introduces a novel method for image local editing, which edits a specific region in the original image with an intuitive instruction. This method allows for precise local editing without affecting other areas of the image, resulting in a realistic final composite image. Malicious groups may exploit this advantage to spread false information or cause misunderstanding. However, we believe that the harm caused by such improper usage can be mitigated with AI-generated content watermarking algorithms or supervising regulations.

\end{document}